\documentclass[11pt]{article}

\usepackage[margin=1in]{geometry}
\usepackage{microtype}
\usepackage{fancyhdr}
\usepackage{graphicx,amsmath,amssymb,amsfonts,amsthm,mathrsfs,bm}
\usepackage{booktabs,multirow,makecell,array}
\usepackage{algorithm,algpseudocode,listings}
\usepackage[dvipsnames]{xcolor}
\usepackage[numbers,sort&compress]{natbib}
\usepackage[T1]{fontenc}
\usepackage[utf8]{inputenc}
\usepackage{hyperref}

\begin{document}


\begin{center}
\rule{\textwidth}{1.5pt}\par\vspace{0.8em}
{\LARGE\scshape A Computationally Efficient Non-Interpolation Virtual Trellis Approach for Mammogram Gradient-based Patterns Analysis\par}

\footnotetext{
\textbf{Preprint Notice:}
This manuscript is a preprint version of a research article currently under submission for publication in a peer-reviewed journal. It has not yet been formally
accepted for publication.}




\vspace{0.4em}\rule{\textwidth}{1.5pt}
\vspace{2em}
\begin{minipage}[t]{0.47\textwidth}\centering
\textbf{Shahid Rasheed}\\
School of Information Technology\\
Deakin University, Australia\\
\texttt{\small s.rasheed@deakin.edu.au}
\end{minipage}\hfill
\begin{minipage}[t]{0.47\textwidth}\centering
\textbf{Muhammad Rashid}\\
Department of Computer Science\\
University of Turin, Italy\\
\texttt{\small muhammad.rashid@unito.it}
\end{minipage}
\vspace{1.5em}
\begin{minipage}[t]{0.47\textwidth}\centering
\textbf{Seifedine Kadry}\\
Department of Applied Data Science\\
Noroff University College, Norway\\
\texttt{\small skadry@gmail.com}
\end{minipage}\hfill
\begin{minipage}[t]{0.47\textwidth}\centering
\textbf{Salman A. AlQahtani}\\
Department of Computer Engineering\\
King Saud University, Saudi Arabia\\
\texttt{\small salmanq@ksu.edu.sa}
\end{minipage}
\end{center}
\vspace{2em}
\begin{center}\large\bfseries Abstract\end{center}
\begin{quote}
Feature analysis is a crucial step in medical image analysis, particularly in breast cancer classification. Traditional image analysis methods often struggle with optimal patterns along the diagonal axis, computational efficiency and feature buffer storage. Different classical operators require high-efficiency hardware for the complex operations, such as image convolution and gradient analysis, which are used for processing complex machine learning models for training and test operations. Here, in this research, we introduce a non-interpolation virtual trellis-based (NVHT) operator that utilizes a filter kernel over the same square image without transformation to a hexagonal trellis. This proposed method is very efficient in reducing computational and storage costs. The proposed operator is used for improved feature extraction efficiency and accuracy in disease classification. This method outperforms conventional approaches by leveraging a hexagonal pixel representation while maintaining high sensitivity in different directions with less storage.
\end{quote}
\vspace{1.3em}

\section{Introduction}\label{sec1}

The computer vision field is rapidly improving the image analysis methods \cite{zhang2024challenges}\cite{dong2024survey} and is set to revolutionize the healthcare field. The techniques and technology help in designing powerful artificial intelligence (AI) algorithms \cite{adeyeye2024artificial}\cite{han2024optimizing} for optimal image acquisition \cite{zhang2024image} through sensors and cameras. As an output, computer vision and AI contribute toward the efficient and quick identification \cite{dang2024computer} and classification \cite{upadhyay2025deep}. It results in accurate and fast diagnosis, recommended systems \cite{arakeri2024intelligent} for monitoring and medication used for better health care outcomes. Sample medical image processing with different modality \cite{darzi2024review}\cite{iniyan2024enhanced} for breast cancer screening \cite{thakur2024systematic}\cite{gabriela2025exploring} \cite{sushanki2024review}.

Breast cancer is initially be examined by a patient manually, or with an annual examination by the doctor. During this examination, doctors advise the scan/mammogram for detailed visualization of the disease \ if it\cite{velayuthapandian2024mammogram} exists. All this is made possible with the help of advancements in technology and digital image processing methods \cite{narumalani2024digital} in computer vision and AI. With the help of computer vision, medical solutions become more optimized and accurate. Computer vision not only tends to increase computational power \cite{alahmari2025comprehensive}  with the help of GPUs and TPUs, but also provides a more efficient framework to handle, train, test, and validate the annotated data \cite{klie2024analyzing}. In the medical domain, cancer data related to oncological scans \cite{xia2024deep}\cite{cook2025impact} and radiological imaging \cite{paverd2024radiology}  are now treated effectively with the help of computer vision. Nowadays, cancer like breast cancer \cite{kaffashbashi2024augmented} is efficiently visualized and treated with the help of computer vision and digital image acquisition devices.
Breast cancer remains a leading cause of mortality among women worldwide. Early and accurate detection of breast cancer lesions is crucial for effective treatment and improved survival rates. Breast cancer detection relies on accurate medical imaging analysis, where edge detection \cite{zhu2024early} plays a key role in segmenting and identifying abnormal growths. Traditional imaging modalities [10], such as mammography and ultrasound, often rely on rectangular pixel structured devices \cite{mireles2025widefield} during image representation. The devices are based on square trellis structure when the image is acquired on display devices.

\begin{figure}
\centering
\includegraphics[width=0.8\linewidth]{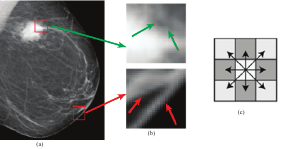}
\caption{Breast cancer mammogram image analysis: (a) original image; (b) ringing (green arrow) /aliasing effect (red arrow) (c) square grid generating these problems during image processing.}
\label{fig:fig_2}
\end{figure}

Conventional devices are now facing problems of aliasing effect with a staircase effect near the affected area. This effect is due to the square-structured image acquisition devices based on the square-structured model in camera, as shown in Figure \ref{fig:fig_2} part (d), which are used for boundary and edge detection \cite{ajlouni2024enhancing} during digital examination. Traditional edge detection techniques \cite{rahman2024nol}\cite{al2025comparison} like Sobel, Canny, and Prewitt frequently have issues with structural accuracy and noise handling. On the other hand, edge detection methods based on hexagonal trellises have become a viable substitute, providing improved spatial resolution and a more realistic depiction of circular structures. Because of its better connectivity and fewer aliasing artifacts \cite{fan2024aliasing}, the hexagonal pixel representation has been investigated in a number of fields. 

\begin{figure}
\centering
\includegraphics[width=0.9\linewidth]{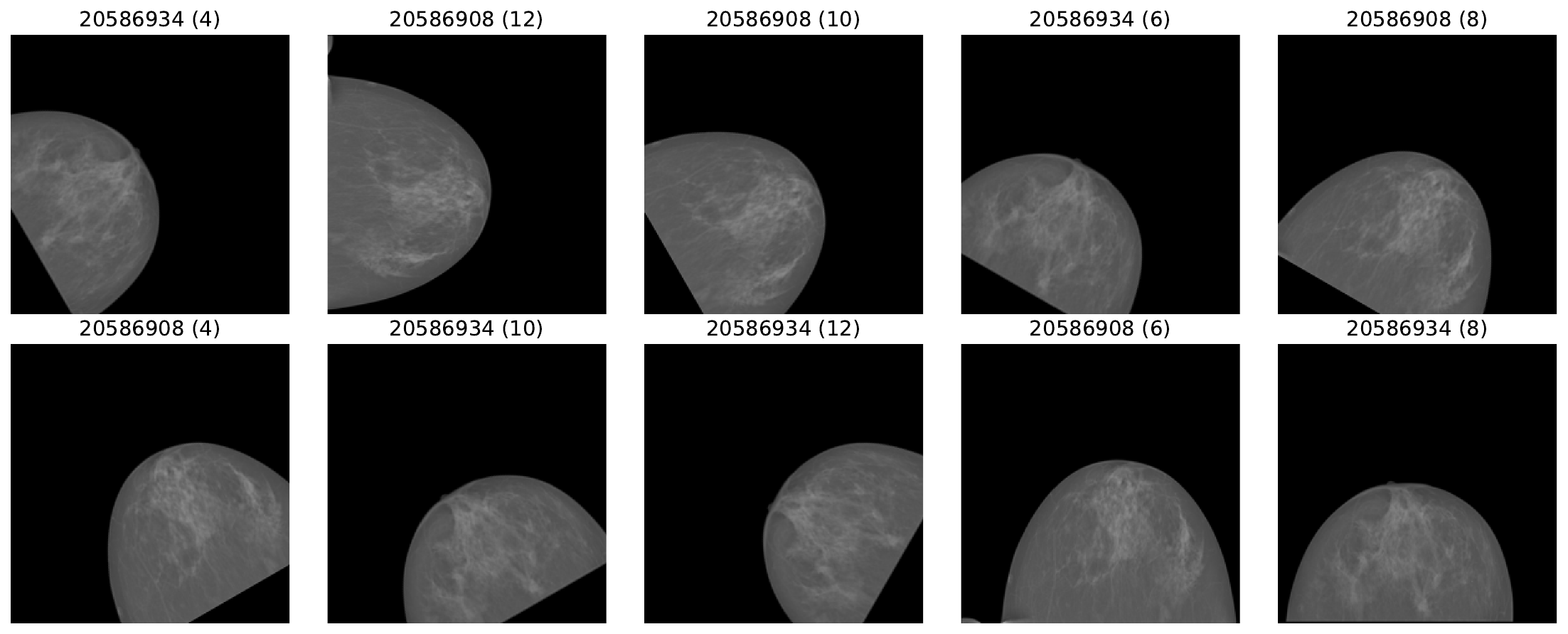}
\caption{Raw images collected from the CHSJ dataset at different annotations \cite{moreira2012inbreast} }

\label{fig:fig_3}
\end{figure}

This study suggests using an optimized filter kernel \cite{zhang2025gaussian} in conjunction with a virtual hexagonal trellis \cite{gardiner2024enhancing}\cite{nethravathi2025multiclass} for image feature analysis \cite{guo2024novel} in breast cancer imaging. The INbreast database\cite{moreira2012inbreast} Figure \ref{fig:fig_3}  contains the breast cancer dataset images that was initially gathered from the Central Hospital de S. Joao (CHSJ).

There are two types of breast cancer in the CHSJ dataset. There are about 5000 grayscale mammograms in one class that are classified as malignant cancer and 2520 images that are classified as benign cancer. We selected this dataset since the raw photos are grayscale and do not require conversion to other color channels, and it is mostly used for breast cancer classification \cite{domingues2012inbreast}. Second, the suggested VHT filter's \cite{rasheed2022single} recognition of the image patterns and clarity are essential to the validation of the suggested method. Preprocessing of the initial CHSJ dataset images is necessary to avoid additional dark regions and enhance the ROI.

\section{Related Work}\label{sec2}
Over the years, several traditional techniques have been applied to different medical imaging modalities with varying degrees of success. In \cite{triwibowo2023identification}, the application of Canny edge detection to ultrasound images achieved a high edge detection and classification accuracy of 95\% each. This method's noise-resilient multi-stage algorithm is renowned for providing the best edge detection and localization. Sobel edge detection [24] was applied to mammography pictures, resulting in 92\% classification accuracy and 93\% edge detection.  Because of its ease of use and effectiveness in identifying edges based on gradient magnitude, the Sobel operator is highly valued. The edge detection applied over MRI images \cite{xu2024advances}, Prewitt edge detection, a close variation of Sobel, slightly increased edge detection accuracy to 94\% and classification to 93\%, demonstrating its suitability for detecting subtle structural alterations in soft tissues.  

\begin{figure*}
\centering
\includegraphics[width=0.8\linewidth]{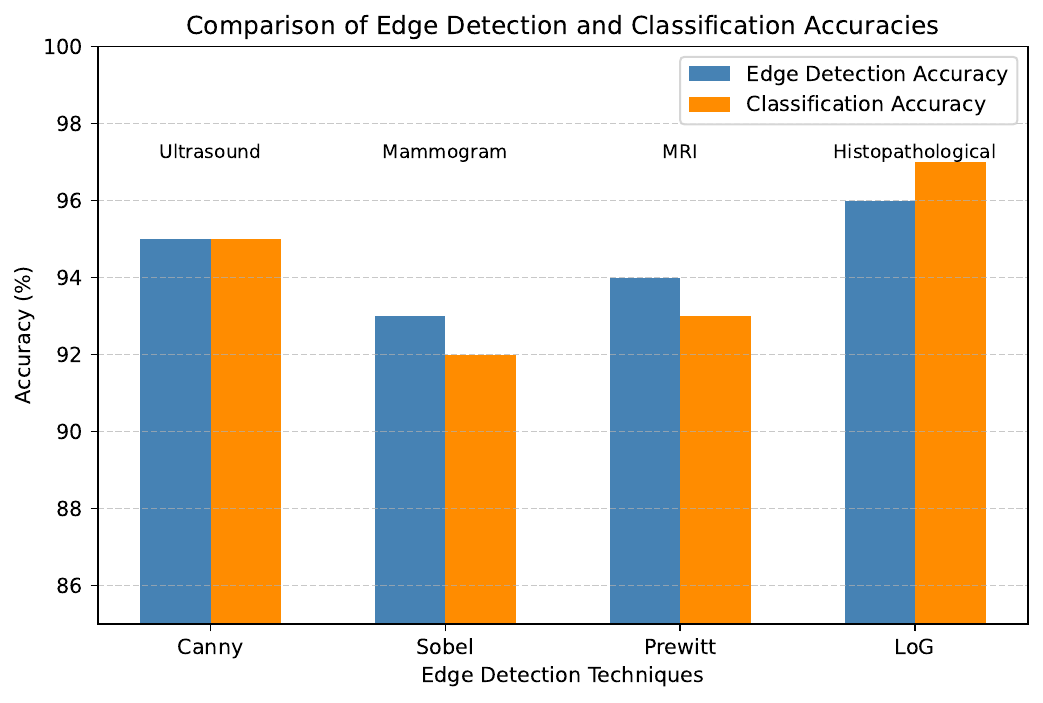}
\caption{Classical filter-based methods used for breast cancer using different datasets }

\label{fig:fig_4}
\end{figure*}

The application of the LoG approach \cite{liqin1994edge}\cite{gurcan2009histopathological} in edge detection and classification in histopathology image data. It has received much attention due to its 96\% edge detection accuracy and 97\% classification accuracy. LoG is helpful in complex boundary identification, as it applies Gaussian smoothing followed by second-order derivative processes, which are often required in histopathology data. Cumulatively, these studies reflect the progression in edge detection algorithms and their application in different branches of medical imaging and their importance in accurate diagnostic tools.
This review focuses on different research contributions that have applied square trellis-based edge detection algorithms for the detection of breast cancer lesions. The analysis is based on the year of publication, type of image, edge detection technique applied, and other relevant performance measures. It provides an extensive summary of the development in the area, which is presented in Table~\ref{tab:tab1}, and Table~\ref{tab:tab2}. 

\begin{table}[htbp]
\centering
\caption{Image intensity-based gradient effect over square trellis
\label{tab:tab1}}

\begin{tabular}{|c|p{3.0cm}|p{2.0cm}|p{2.0cm}|p{3.5cm}|}
\hline
\textbf{Cite} & \textbf{Technique} & \textbf{Features Extracted} & \textbf{Image Type} & \textbf{Remarks} \\
\hline
\cite{gao2025segmented} & Square-Triangle Tessellation Descriptor (STTD) & Not specified & MPEG-7-PartB, Tari-1000, Kimia's 99 & Captures shape geometry for efficient shape retrieval using a square trellis. \\
\hline
\cite{latif2024discrete} & Discrete Wavelet Transform (DWT) & Not specified & General image datasets & Multi-resolution edge detection using frequency decomposition. \\
\hline
\cite{zhao2015enhanced} & Histogram of Oriented Gradients (HOG) & $\sim$378 features per 64$\times$128 image & Pedestrian detection datasets & Square grid cells used to compute gradient orientation histograms. \\
\hline
\cite{kim2020study} & Canny Edge Detection with square grid sampling & Not specified & General image datasets & Localizes edges using intensity gradients and non-maximum suppression. \\
\hline
\end{tabular}
\end{table}
\begin{table}[htbp]
\centering
\caption{Deep learning methods for gradient effect detection with key findings. 
\label{tab:tab2}}
\begin{tabular}{|c|p{1.6cm}|p{1.9cm}|p{1.8cm}|p{5.0cm}|}

\hline
\textbf{Year} & \textbf{Image Type} & \textbf{Edge Detection Technique} & \textbf{Performance Metrics} & \textbf{Key Findings} \\
\hline
\cite{dey2022screening} & Thermogram & Deep Learning & Classification accuracy & Integrated edge detection with deep learning for accurate lesion classification. \\
\hline
\cite{mostafa2015edge} & Thermogram & Morphology on Hexagonal Grid & Edge enhancement & Proposed a method for edge enhancement using morphological operations on hexagonal grids. \\
\hline
\cite{mostafa2015edge} & Mammogram & Binary Morphology & Edge detection efficiency & Demonstrated superior edge detection on hexagonal grids compared to rectangular grids. \\
\hline
\cite{nabizadeh2015brain} & MRI & Gabor Wavelet & Edge detection accuracy & Found that Gabor wavelets on hexagonal grids outperform rectangular grids in MRI images. \\
\hline
\end{tabular}
\end{table}

The above line and bar chart visualize the classification accuracy of various hybrid edge detection and machine learning models applied to medical imaging from 2015 to 2024. These methods primarily enhance early detection of breast cancer through preprocessing techniques using square and hexagonal trellis structures.


Integrating edge detection with machine learning algorithms, a hybrid model \cite{al2024advanced} achieved a classification accuracy of 95.3\% . This approach, however, was still expensive in terms of computation and data. Among all techniques considered, hexagonal trellis frameworks maintained a consistent edge \cite{rashid2023white}. These frameworks offered spatial resolution and uniform sampling that were superior to traditional square trellis methods, thus yielding better edge and classification precision. Despite these advantages, there are hurdles to overcome, such as the need for efficient conversion algorithms for images captured on rectangular grids and hexagonal grids, which may be more computationally expensive due to complex algorithms. Many researchers have focused on edge detection for medical images. Traditional edge detection techniques have weaknesses; for example, the Sobel edge detector uses a gradient-based convolution \cite{saif2016gradient}, which is sensitive to noise. The Canny edge detector is more computationally expensive \cite{zhang2024challenges} due to its multi-stage filtering approach, and wavelet-based methods are difficult because of their complex implementation despite multi-scale edge detection. 
New methods in the processing of hexagonal images have shown great potential for spatial efficiency and connectivity, and have motivated the need for a Virtual Hexagonal Trellis-based approach in medical imaging. This method changes the processing technique and also imparts optimization effects in the existing methods based on low computational cost and less storage. We only handle the transformation of the image, which may cause loss of information during transformation. For all these factors, we proposed a hexagonal trellis method with a non-interpolation technique to avoid information loss and gain high accuracy in pattern detection in all gradient directions.

\section{Proposed Methodology}\label{sec3}

\begin{figure*}
\centering
\includegraphics[width=0.9\linewidth]{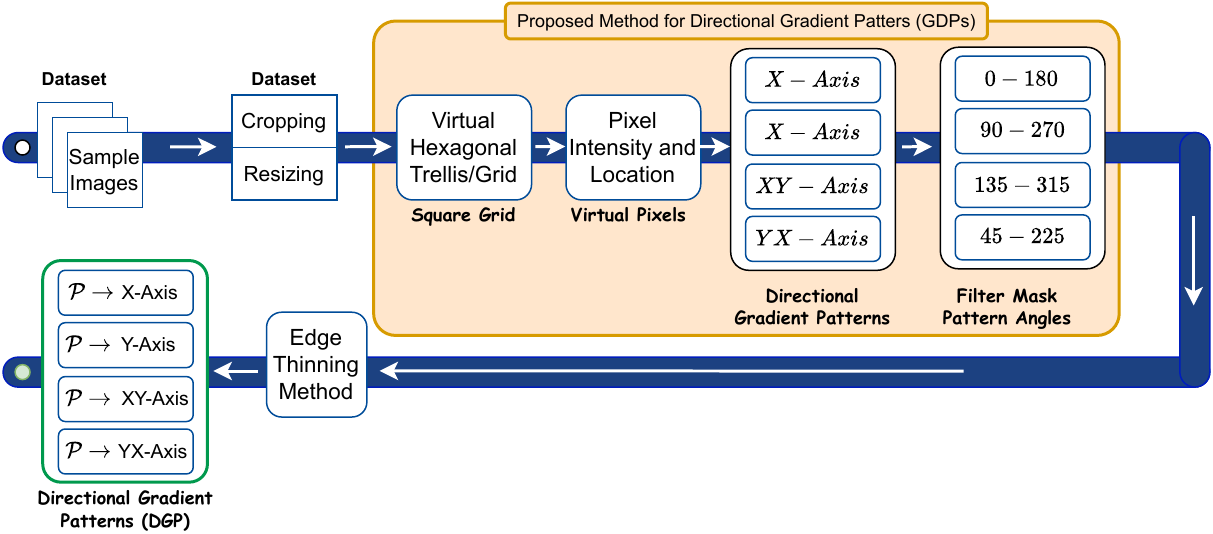}
\caption{Proposed Method Block Diagram}
\label{fig:fig_proposed}
\end{figure*}

Classical image processing methods are not only based on a square trellis for image acquisition but also represent a poor representation of the direction gradient effect during edge /boundary detection in medical imaging, such as breast cancer mammograms, CT, and other MRI imaging modalities. For the effective handling of information, delicate images are converted to a hexagonal trellis, which is a method for the representation of complicated details. But still, it still costs computational resources and requires high storage during transformation and quantization from square to a hexagonal trellis. To overcome these problems, we proposed a hexagonal trellis structure over the same square trellis virtually, by avoiding the overhead of image transformation and reducing image storage, which shows strong information preservation during the image processing. A block diagram of the proposed method is presented in Figure \ref{fig:fig_proposed}

\subsection{Dataset and Preprocessing}
\begin{figure*}
\centering
\includegraphics[width=0.6\linewidth]{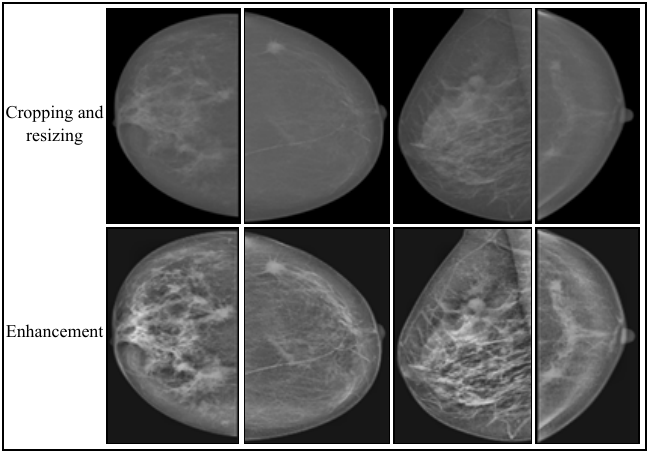}
\caption{Preprocessing CHSJ breast cancer dataset images}
\label{fig:fig_7}
\end{figure*}

Breast cancer dataset CHSJ, which is publicly available at Zenodo, published by Universiti Kebangsaan Malaysia on 30th january 2025 \cite{hussain_2025_14769221}. This dataset two categories of breast cancer mammograms in the dataset: benign and malignant. This dataset consists of mammograms obtained from 100 subjects, and about 72 or more images are taken from various projection angles for each subject. In both acquisitions, both raw and refined images are taken into account, where the refined image is kept next to its corresponding raw image. There are a total of 7,632 images in this complete dataset, including raw and refined images in both categories. The sample images are shown in Figure \ref{fig:fig_7}. The images contain extra dark areas and noise in the image. To refine the images for the proposed method implementation, we preprocessed the images using resizing and normalization.

\subsection{Virtual Hexagonal Trellis}
\begin{figure*}
\centering
\includegraphics[width=1\linewidth]{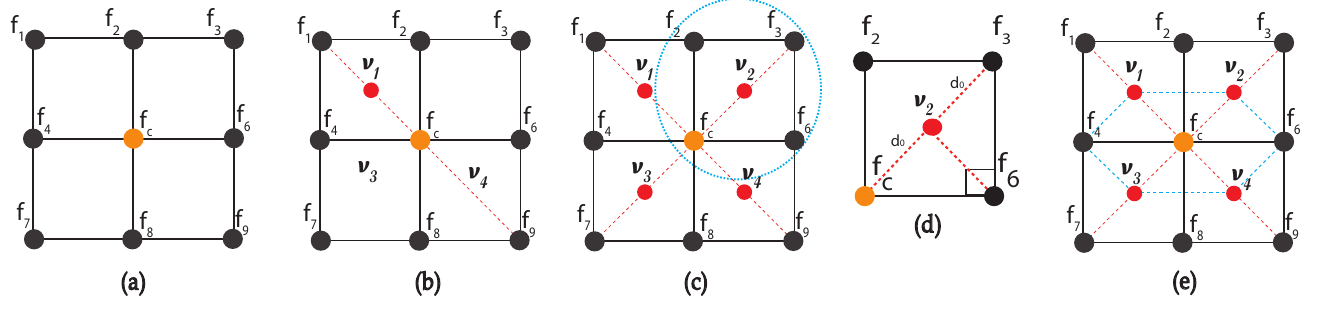}
\caption{Proposed Virtual Trellis for pseudo virtual patterns (PVP) detection operators: (a) Sample square trellis (b) pseudo virtual pixel (PVP) generation (c) location and intensity of PVP; (d) PVP half-distance and intensity calculation from the central pixel (e) PVP replacing four neighboring pixels 
}
\label{fig:fig_8}
\end{figure*}

The hexagonal trellis grid is virtually implemented over a \textit{Cartesian grid} by redefining virtual pixels at half distance with neighboring real pixels.  This approach enhances edge continuity and reduces jagged boundaries common in square grid processing. The core computing concept in the proposed techniques is dependent over the virtual pixels and the kernel selection and derivation process. As we derived kernels based on the hexagonal textural patterns whenever the filter convolves over the image. The basic derivation and trellis concept of NIM is presented in Figure \ref{fig:fig_8}.

\subsection{Virtual pixel intensity and location}

The entire virtual hexagonal trellis is based in the method of \textit{PVPs} generation. When the \textit{PVPs} are generated between the neighboring and central pixel, the aliasing effect reduces. To visualize the effect the virtual pixels are assigned the average intensity of the relevant neighboring and central pixels. The second most important aspect of the \textit{PVPs} is the location where they are placed. The \textit{PVPs} are placed at $ \frac{d_o}{2}$ distance as shown in figure \ref{fig:fig_8}(d). Similarly, four neighboring pixels are replaced with the new virtually generated \textit{PVPs} with new intensity and location. Using new, virtually generated hexagonal trellis over the same rectangular grid, we derived a hexagonal filter mask with four directional gradient operators for pattern recognition.\\
\subsection{Hexagonal Filter Kernel}
A specialized convolution hexagonal filter kernel (\textit{CHFK}) is designed to operate by reflecting the hexagonal trellis framework. The kernel is convoluted for directional gradients for better edge localization. These \textit{DGPs} at first stage shows the localization of edges from $0^{\circ} - 180^{\circ}$  along positive x-axis to negative x-axis. 

\begin{figure*}
\centering
\includegraphics[width=0.4\linewidth]{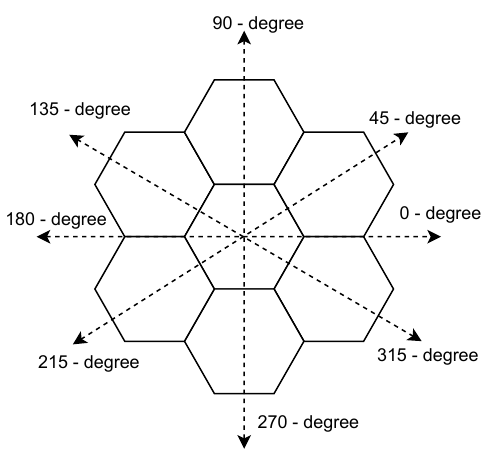}
\caption{Intensities and angle of direction gradient patterns}
\label{fig:fig_9}
\end{figure*}

The second stage of CHFK reflects the localization of high intensity pixels from (90° - 270°) along the positive y-axis to negative y-axis. These two stages result in the DGPs from x-axis to y-axis. These stages result in optimized edge detection with optimized boundary or shape feature as compared to the square trellis DGPs. Further, optimization is incorporated in the proposed method by introducing two other stages of CHFK along the xy-axis and yx-axis. The 3rd stage of CHFK extracts the edges with more optimized DGPs from (45° to 225°), finally the 4th stage of CHFK convolution shows the DGEs from (135° to 315°) in the Figure \ref{fig:fig_9}.

\begin{figure*}
\centering
\includegraphics[width=0.9\linewidth]{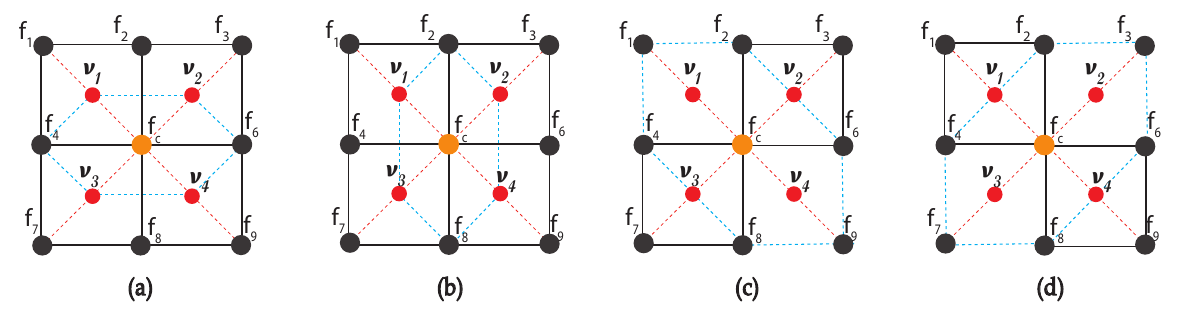}
\caption{Intensities and angle of direction gradient patterns (a) x-directional gradient operator (0$^{\circ}$ - 180$^{\circ}$) (b) y-directional gradient operator (90$^{\circ}$ - 270$^{\circ}$) (c) yx-directional gradient operator (135$^{\circ}$ - 315$^{\circ}$), and (d) xy-directional gradient operator (45$^{\circ}$ - 215$^{\circ}$) 
} 
\label{fig:fig_10}
\end{figure*}

These derivation results in a variety of different variations of the hexagonal trellis-based kernels and we presented only one variation at different directional gradient patterns (DGPs), which shows the optimal results in terms of edge detection with reduced jogging effect and with high computational efficiency. These all-directional gradient patterns with all pseudo angles and intensity directions are represented in Figure \ref{fig:fig_10}. First, the x-DGOs are represented with directional angles which are at 0° - 180°, similarly, y-DGOs are at 90° - 270°, xy–DGOs at 45° to 225°, and finally, the yx–DGOs are at 135° to 315°. These four variants are implemented on the breast cancer mammographs for pattern recognition to optimize the feature extraction methods using a novel hexagonal trellis.

Due to the non-transformation from square to hexagonal trellis, the method also reflected the non-interpolation method for hexagonal image processing (NIM-HIP). With this novelty of NIM, in near future, the same method will be a part of the first hexagonal image acquisition device in the field of computer vision. Figure  \ref{fig:fig_10} shows the four different DGPs derived from the hexagonal trellis operation. On the left, DGPs along x and y-axis and on right, the DGPs of left and right diagonal along with the convolutional hexagonal operator kernel for edge detection.

\subsection{ Implementation and Workflow }

\begin{figure*}
\centering
\includegraphics[width=1\linewidth]{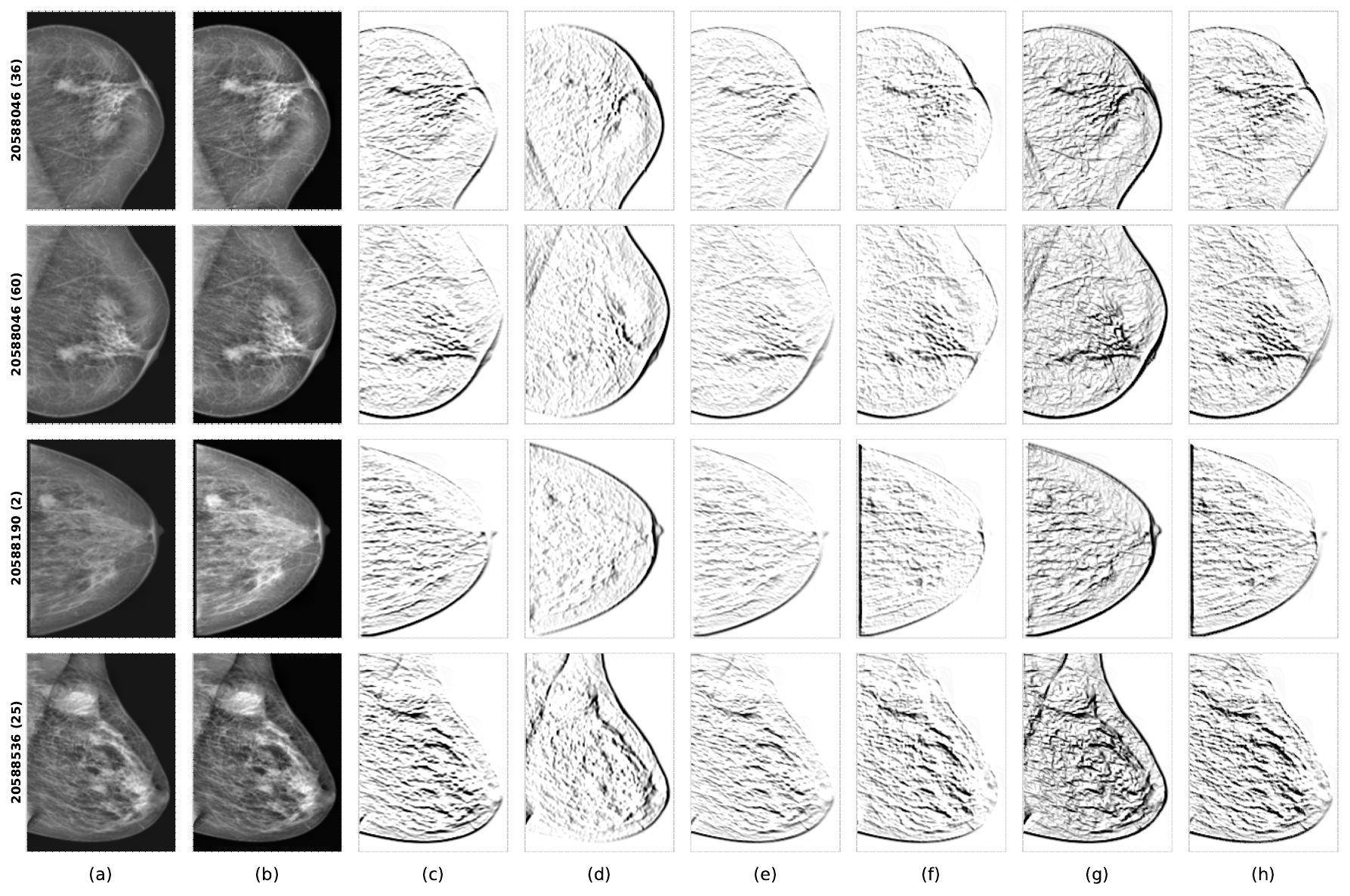}
\caption{Direction gradient patterns along different axis and angle using VHT operator, (a) Original image (b) grayscale channel (c) DGP along x-axis (d) DGP along y-axis (e) DGP along xy-axis, (f) DGPs along yx-axis, (g) combined results of DGPs along x an d y, (h) and finally, combined results of DGPs along x and y.}
\label{fig:fig_11}
\end{figure*}

The prepressed normalized breast cancer mammogram images are evaluated for the DGPs using VHT filter variations. The first directional gradient operator along the x-axis is convoluted over the images to get the patterns whose gradient intensity direction is aligned with x-axis either in positive or negative direction e.g. (0$^{\circ}$ - 180$^{\circ}$). Finally, the combined DGPs effect along diagonal xy (45$^{\circ}$) and yx (135$^{\circ}$) are also shown for analysis, with colum labeled as combined\_xy'' and “combined\_yx). This operator collects the patterns having intensity scattering along the positive and negative x-axis. This operator not only extracts DGPs but also a smooth scattering effect as shown in Figure \ref{fig:fig_11}. This smooth scattering is not collected in the classical filters when used for edge \/ boundary feature analysis .

Similarly, in the proposed Virtual Hexagonal Mapping, we applied the hexagonal trellis overlay to restructure pixel relationships. This structure extracts the patterns along y-axis (90° - 270°), along xy-axis (45° - 225°) which we called the diagonal direction, and finally the yx – axis (135° - 335°) also reflected diagonal direction. By implementing the proposed VHT based DGPs we extract almost all directional patterns from 0° to 360°. On the other hand, the already traditional methods only extract the patterns in x-axis and y-axis with an angle of 0° - 180° and 90° - 270° respectively. The visualization and the extracted patterns with respect to the VHT-DGPs variation.

From the  Figure \ref{fig:fig_11}, we can analyze that the exact patterns using VHT operator variations are very thick and coinciding with the other neighboring patterns, which make the resultant image unclear about the certain patterns. If the cancer area lies in the same coinciding patterns, then it is very difficult to evaluate the patterns. To overcome this issue, we use a next stage post processing to refine the patterns.

\subsection{Post-processing}
Apply the edge / future thinning and morphological operations for refined these patterns. The extracted patterns are thick and mostly considered with the other patterns. The Canny edge detection algorithm uses an initial input of 80 and 200 thresholds to output a binary edge map and image contour. It tracks continuous edges and detects rapid intensity changes. The image is converted to binary by setting edge pixels to 1 and non-edge pixels to 0, which is beneficial for thinning edges. This process reduces the image to a single pixel while maintaining topological structures and eliminating redundant edge thickness. 

\begin{figure*}
\centering
\includegraphics[width=0.8\linewidth]{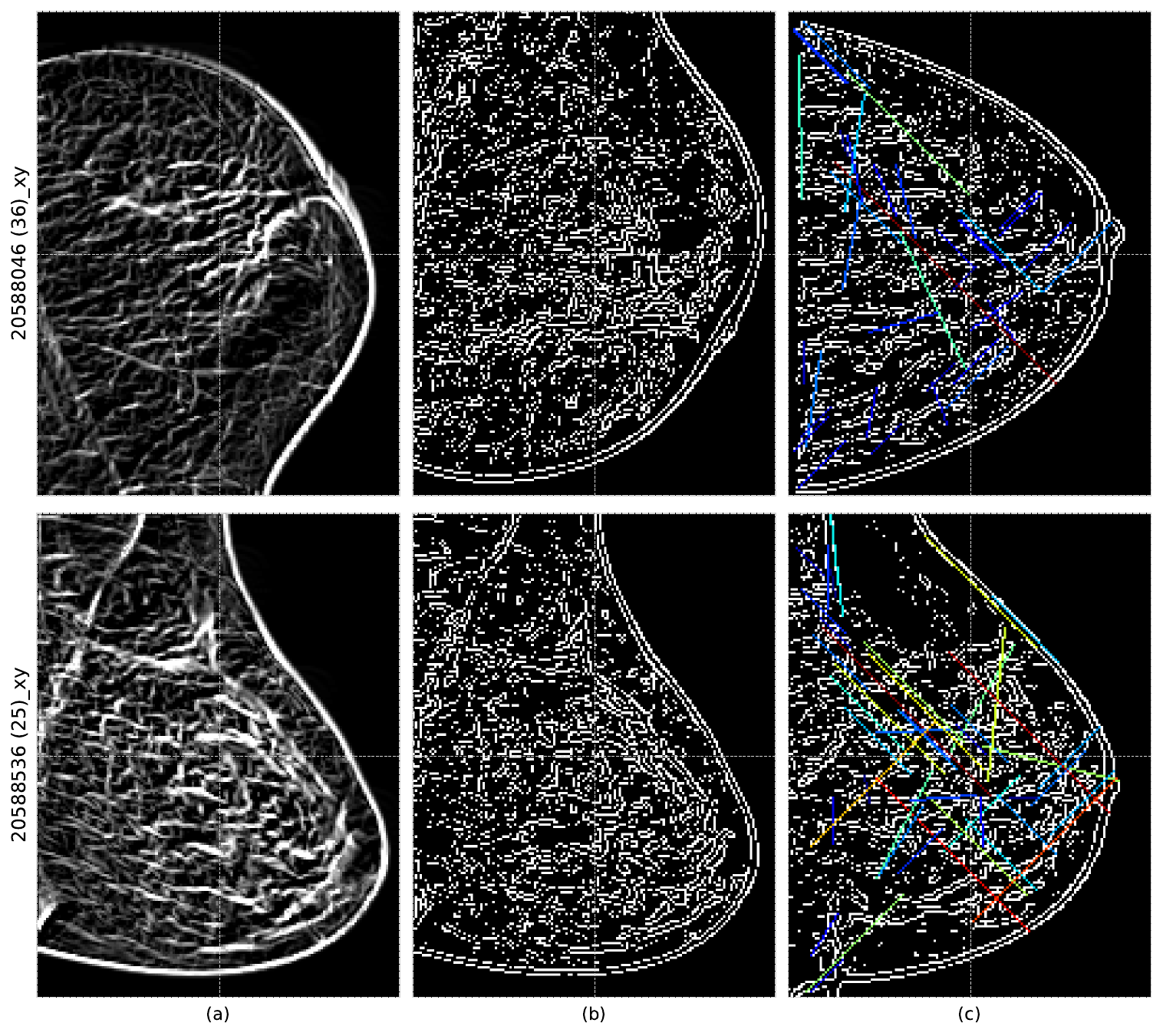}
\caption{Pattens thinning using post processing (a) NIVT processed thick edges (b) NIVT post processed thin edges (c) color labeled connected edges }
\label{fig:fig_12}
\end{figure*}

Standard OpenCV functions are only applicable to formatted images. The image is then converted back to skeletonized form to uint8, and directional edges are processed through contour thinning and refinement. Edge filters are convolved to create black backgrounds and white images. To rectify this issue, we process the resultant images with edge / patterns thinning in post processing of the images. The post processing method as presented in figure 12, not only gave the think pattens but also a reflection of the structured original image with very light weight gradient effects as shown in figure 13, for any future processing in the field of computer vision.

\begin{figure*}
\centering
\includegraphics[width=0.7\linewidth]{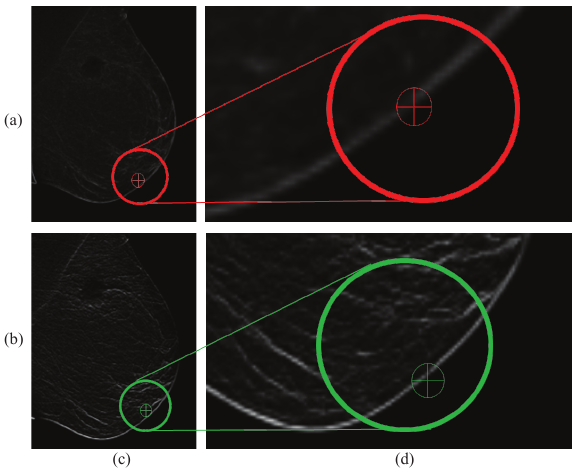}
\caption{Figure 13. Comparative analysis of the traditional square trellis with the proposed NIVT on the top left and bottom of the image. (a) (c) Square trellis shows a blurry and low intensity edge with aliasing effect (b) \& (c) Proposed NIVT  method, represents the high intensities on the edges and better pattern visualization. (a) (d) images represent the dim and blurry edges when zoomed at 30x, and (b)(d) clearly show crisp edges and dominate over the square trellis.}
\label{fig:fig_13}
\end{figure*}


\subsection{Performance Metrics}

Evaluation metrics and parameters used for validation and quantification of proposed methods are edge count, EPI, and entropy for internal information analysis based on the intensity of the resultant image. One other aspect is computational time, which is also included to estimate the processing time of both the square trellis-based method and the proposed VHT-based method. Computational time is important in medical imaging, especially where large-scale remote storage is used to store and access large amounts of data. All these parameters are evaluated in both existing edge or pattern extraction methods like Sobel, Robert, and Canny detectors and proposed NIVT method.

\section{Results and Discussion}
\begin{figure*}
\centering
\includegraphics[width=0.99\linewidth]{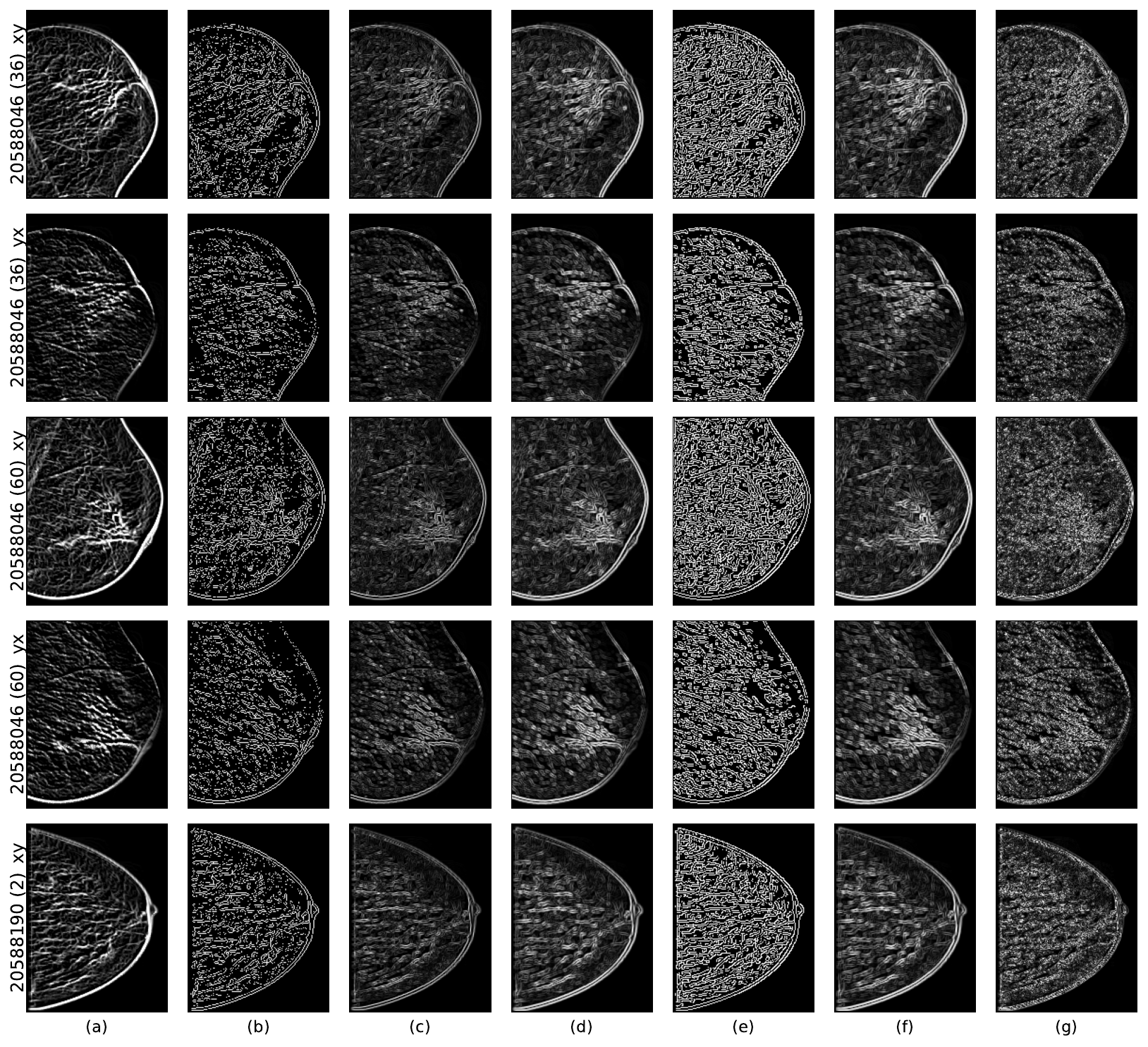}
\caption{Comparative analysis of the proposed method with different classical filters for edge pattern detection (a) NIVT edge detected (b) NIVT* (c) Robert (d) Sobel (e) Canny (f) Prewitt (g) Laplacian }
\label{fig:fig_14}
\end{figure*}

The performance of conventional edge detection operators (Robert, Sobel, Canny, Prewitt, and Laplacian) was compared against the proposed NIVT (Novel Integrated Virtual Hexagonal Trellis) method. The qualitative results in comparison with classical methods are shown in Figure \ref{fig:fig_14}. 

The other parametric evaluation was conducted using multiple quantitative parameters, including the number of detected edges, edge preservation index (EPI), connected edges, and entropy Table 3.  Overall, all three techniques are fairly distinct in their edge detection and the number of edges. The Canny operator (53,630), Sobel (28,915), and Prewitt (27,919) had the highest number of edges detected with $\pm 10$ to $\pm 30$ maximum and an average variance of $\pm 22$. Meanwhile, Laplacian (23,156) and Robert (23,053) had lower edge detection results with $\pm 10$ to $\pm 30$ variance. The proposed NIVT edge detector has a well-balanced number of edges (29,944 edges) detected relative to the output of the other edge detectors, with $\pm 27$ variance as per different images with different edges and intensity values. The NIVT edge detector output was neither under-edged like Robert and Laplacian, nor over-edged patterns like Canny. This overall balance suggests that the NIVT edge detector contains the features of structural detail that were retained while eliminating excessive noise, with maximum average edges.


\begin{table*}[!t]
\centering
\setlength{\tabcolsep}{7pt}
\caption{Proposed method evaluation based on parametric analysis.}
\label{tab:tab2_duplicate1}

\begin{tabular*}{\textwidth}{@{\extracolsep{\fill}}p{2.7cm} p{2.0cm} p{1.9cm} p{1.3cm} c@{}}
\toprule
\textbf{Edge Operator} &
\textbf{\# of edges} &
\textbf{EPI} &
\centering\arraybackslash\textbf{Connected Edges} &
\textbf{Entropy} \\
\midrule
Robert        & $23053 \pm 30$ & $0.436 \pm 0.041$ & 11 & 8.4  \\
Sobel         & $28915 \pm 22$ & $0.568 \pm 0.027$ & 28 & 9.4  \\
Canny         & $23630 \pm 18$ & $0.415 \pm 0.056$ & 55 & 6.8  \\
Prewitt       & $27919 \pm 25$ & $0.573 \pm 0.022$ & 49 & 8.38 \\
Laplacian     & $23156 \pm 10$ & $0.468 \pm 0.019$ & 10 & 4.74 \\
Proposed NIVT & $29944 \pm 27$ & $0.999 \pm 0.017$ & 63 & 5.56 \\
\bottomrule
\end{tabular*}
\end{table*}

Results in Table \ref{tab:tab2} are calculated on 100 CHSJ dataset images from 20588046 (36) to 20586986 (42), average results with $\pm $ variance over different images.

In our assessment of how well we maintained features, a useful measurement is the Edge Preservation Index (EPI). Specifically, Sobel (0.568) and Prewitt (0.573) had better EPI from traditional methods with $\pm 0.027$ and $\pm 0.022$ variance respectivly, when compared to Robert (0.436), Canny (0.415), and Laplacian (0.468) which yielded moderately decent EPI with $\pm 0.41$, $\pm 0.056$, and $\pm 0.019$ variance on aveagre of 100 images when pocesssed. However, the proposed NIVT showed near-edge preservation with a significantly better EPI (0.999) with minor fluctuation of $\pm 0.017$ variance.  
The analysis of connected edges further illustrates the robustness of the approach being proposed. Canny produced 55 connected edges, Prewitt 49, and Roberta and Laplacian produced very few connected edges with connectivity of 11 and 10, respectively, indicating fragmented edge maps. On the other hand, NIVT produced the most connected edges of 63, demonstrating that it contributes to articulating interconnectedness in edge structures. This interconnectedness of edges is also visualized with color-connected edges in Figure \ref{fig:fig_12} (c), a crucial part of an edge for further downstream tasks like segmentation and classification. When it comes to providing information, Sobel (9.4), and Prewitt (8.38) had the highest values, indicating more complex and meaningful edge maps. Robert (8.4) and Canny (6.8) had a modest amount, and lastly, Laplacian had the lowest entropy value with (4.74). The proposed method, NIVT, had an entropy of 5.56, which was lower than that of Sobel and Prewitt, but suggests controlled information distribution, because if entropy is more than the average value of the other mehtods then noise may be added, or if the value is lower than the aveage value of the other methods, it mean information is lost during processing. This implies that NIVT may reduce redundancy and noise; this would contribute to edge maps being clearer and easier to interpret, with entropy of 5.56 whih is very near to the average value of all methods i.e., 7.21. 

\begin{figure*}
\centering
\includegraphics[width=0.8\linewidth]{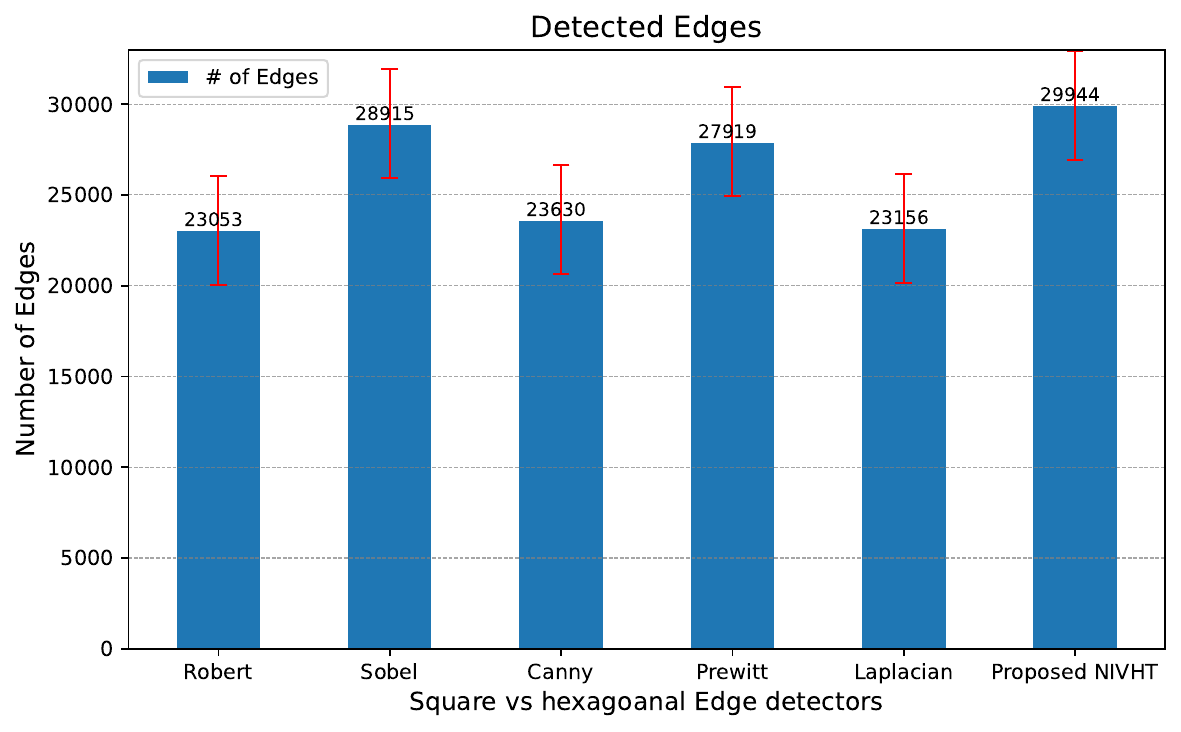}
\caption{Comparative analysis of the proposed NIVT method with other classical methods based on the number of detected edges}
\label{fig:fig_15}
\end{figure*}

\begin{figure*}
\centering
\includegraphics[width=0.8\linewidth]{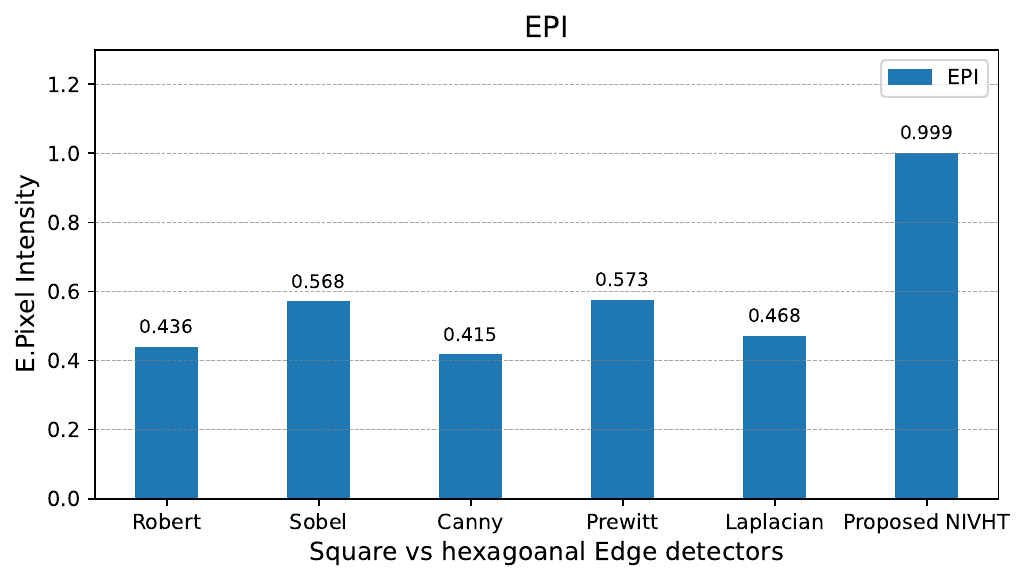}
\caption{Comparative analysis of the proposed NIVT method with other classical methods based on EPI}
\label{fig:fig_16}
\end{figure*}

\begin{figure*}
\centering
\includegraphics[width=0.8\linewidth]{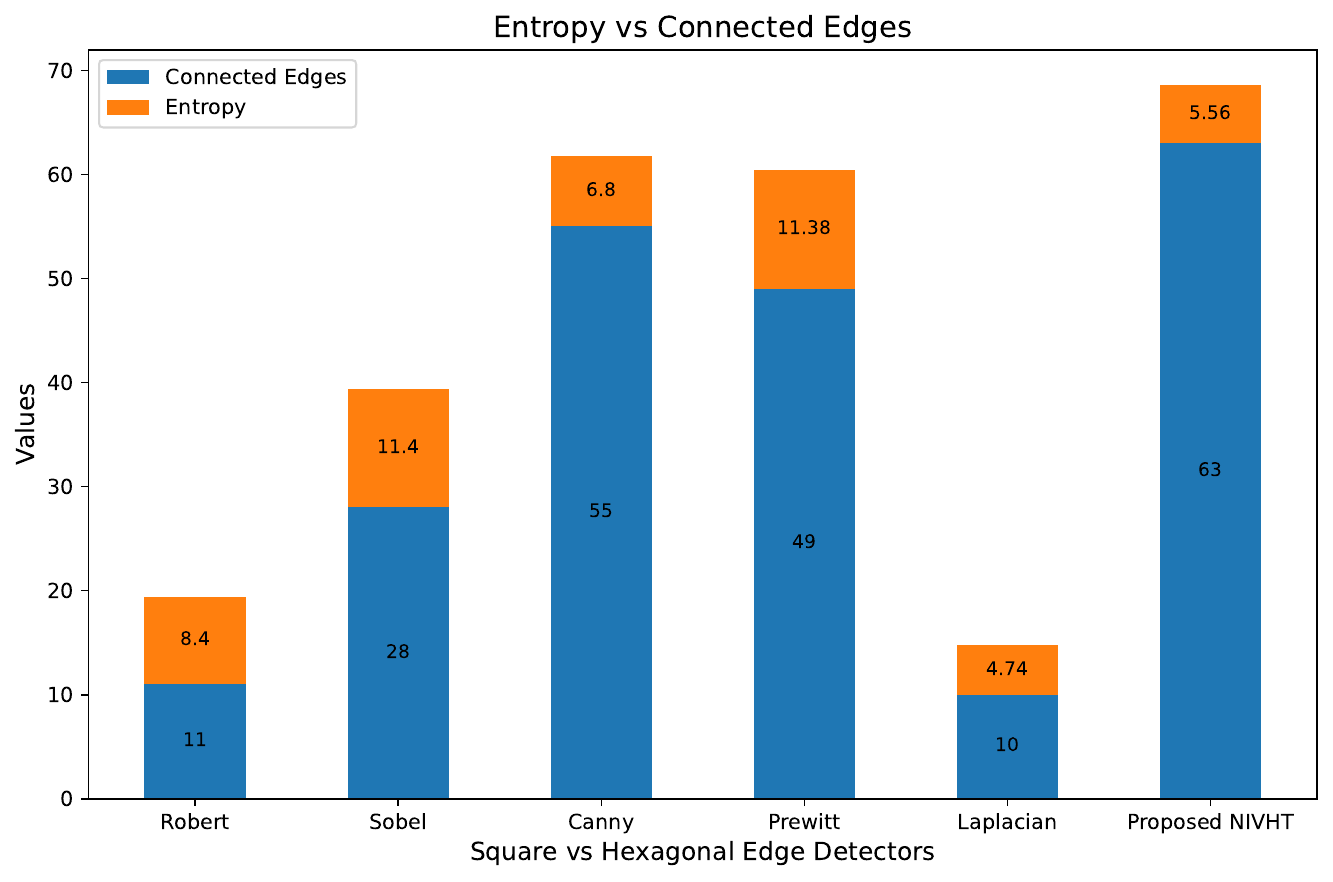}
\caption{Comparative analysis of the proposed NIVT method with other classical methods reflecting the effect of entropy over connected (continuous) edges}
\label{fig:fig_17}
\end{figure*}

The Figure \ref{fig:fig_17} highlights a comparison of connected edges and entropy for each of the edge detection operators, including the newly proposed NIVT method. By presenting the information in these two dimensions, we are able to not only evaluate the structural connectivity of what remains of the detected edges, but also the richness of what edge information remains.

The results show the clear domination of NIVT over Robert and Laplacian operators below others in terms of both parameters. While Robert produced a result of only 11 connected edges with an entropy of 8.4, even worse, Laplacian produced 10 connected edges and the lowest entropy value of 4.74. These results unambiguously depict that structurally continuous edge maps and those with information richness could not be produced using these methods. Both Sobel and Prewitt operators had stronger performance based on their entropy values of 9.4 and 8.38, respectively, indicating richer maps of information content. However, their values of connected edges remained limited at 28 and 49, respectively, suggesting that while richer detail supported the premise, they even had a level of structural connective continuity at least, which was also not optimal at this stage. Similarly, Canny performed relationally, producing 55 connected edges, the second-highest result among conventional methods, but this did not equate to a proportionally good value of entropy, which displayed a low level of 6.8 significance. This highlights that while Canny does promote connectivity of the edges of an object, it also sacrifices considerable levels of detail, which ultimately does reduce the richness of the info, as evidenced by its entropy value being relatively low. The newly proposed NIVT method indisputably surpasses all traditional methods because it achieves 63 connected edges, which is the highest count across all methods, while having a relatively small entropy value of 5.56. The combination of the two indicated that NIVT was producing edge maps that were structurally continuous and had no excessive noise or duplication. 
\\
\begin{figure*}
\centering
\includegraphics[width=1\linewidth]{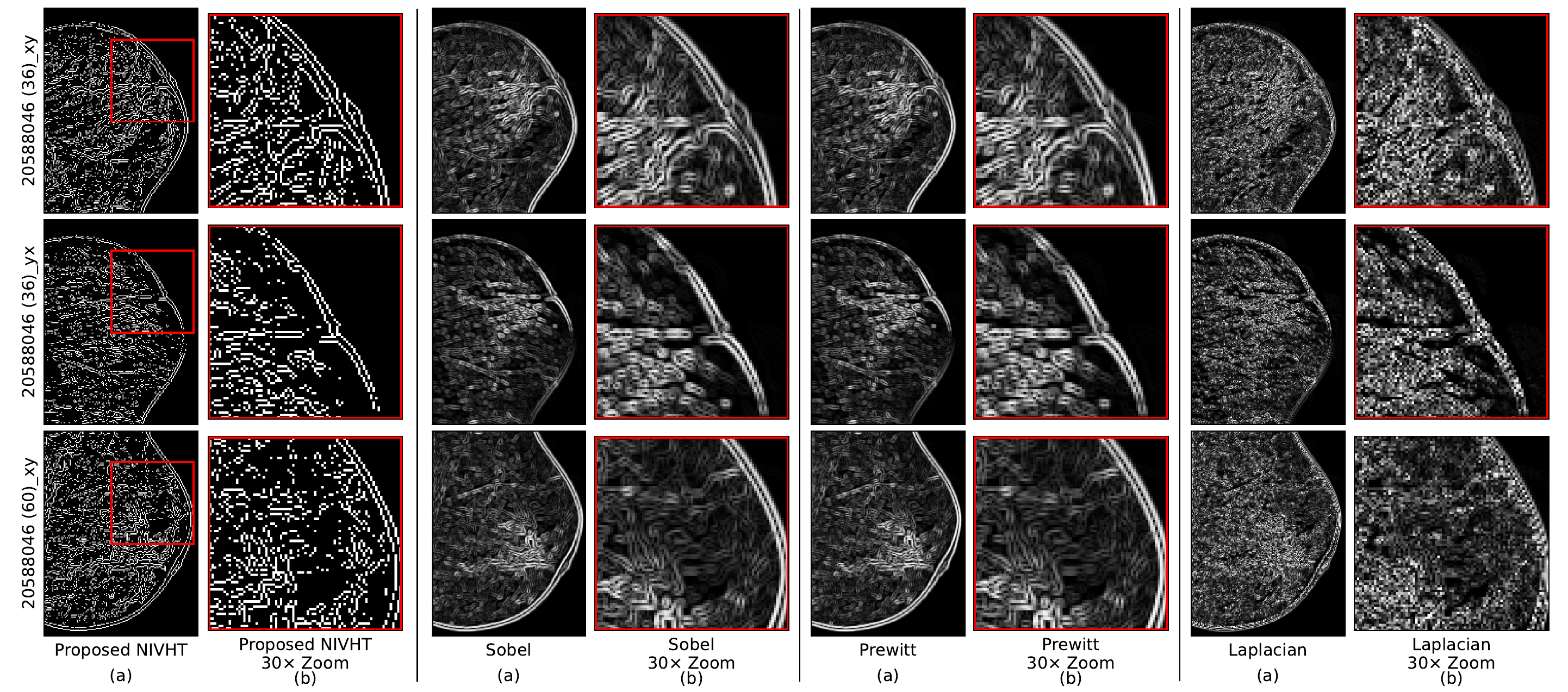}
\caption{Classical method comparative analysis with proposed method w.r.t. boundary/edge structure zoomed visualization (a) classical and proposed methods implemented over the full image (b) 30x zoomed red-square cropped part to better understand the extracted structure from all existing methods}
\label{fig:fig_18}
\end{figure*}
The comparative analysis shows that although operator methods can be superior at certain metrics, they generally seem to trade off richness of edges, connectivity, and preservation of edges. In contrast, NIVT consistently performed across the entire dataset with reasonable amounts of edge detail. The algorithm's high EPI values and large number of connected edges emphasize that NIVT can produce structurally consistent and noise-reduced edge maps, thus providing the basis for further image analysis. 

\subsection{Computational Time Analysis}

We assessed the efficiency of the proposed NIVT method compared to the existing square-to-hexagonal interpolation and reconstruction pipeline by determining the computational time for processing multiple images. In the conventional method, processing takes place in three main stages:

$$Average~time~for~square~trellis~to~hexagonal mapping = \frac{1.5 + 1.3}{2} = 1.4ms $$

$$Average~time~for~hexagonal~Filtering = \frac{2.0 + 5.0}{2} = 3.4ms $$
$$Average~time~for~reconstruction~in~square~trellis = \frac{1.5 + 2.5}{2} = 2.0ms $$
Collectively, the total average processing time in the case of conventional processing can be summarized at approximately 6.8 ms (or 207.6 ms on average when processing 30 images). In contrast, the proposed NIVT method operates as shown in Table 4, at a rate that is consistently about 1.0 ms (or 35 ms in the case of 30 images) because it fits a virtual hexagonal trellis as part of image processing without interpolation, and therefore does not need to perform many of the previous interpolation steps.
\begin{table}[h!]
\centering
\renewcommand{\arraystretch}{1.3}
\caption{Computational cost analysis between the traditional method and the proposed NIVT method}
\setlength{\tabcolsep}{2.0pt} 
\begin{tabular}{|p{1.7cm}|p{1.8cm}|p{2.0cm}|p{2.0cm}|p{1.8cm}|p{1.8cm}|p{1.0cm}|}
\hline
\multirow{2}{*}{\textbf{Image label}} & 
\multicolumn{3}{c|}{\textbf{Transformation and processing time (ms)}} & 
\multirow{2}{*}{\begin{tabular}[c]{@{}c@{}}Square to \\ hex\\ interpolation\end{tabular}} & 
\multirow{2}{*}{\makecell{\textbf{Proposed}\\\textbf{NIVT}}} & 
\multirow{2}{*}{\textbf{Diff.}} \\ \cline{2-4}
 & \textbf{Square to hex mapping} & \textbf{Hexagonal processing} & \textbf{Hex to square reconstruction} &  &  &  \\ \hline
22580098 & 1.40 & 3.4  & 2.0  & 6.80 & 4.016 & 3.984 \\ \hline
20588046 & 1.80 & 1.23 & 1.80 & 4.83 & 0.999 & 3.831 \\ \hline
20588190 & 2.40 & 1.99 & 1.78 & 6.15 & 1.001 & 5.149 \\ \hline
22579730 & 1.10 & 2.01 & 2.22 & 5.31 & 1.202 & 4.108 \\ \hline
22580098 & 1.34 & 0.98 & 1.45 & 3.77 & 0.890 & 2.880 \\ \hline
\end{tabular}
\end{table}

\begin{figure*}
\centering
\includegraphics[width=0.8\linewidth]{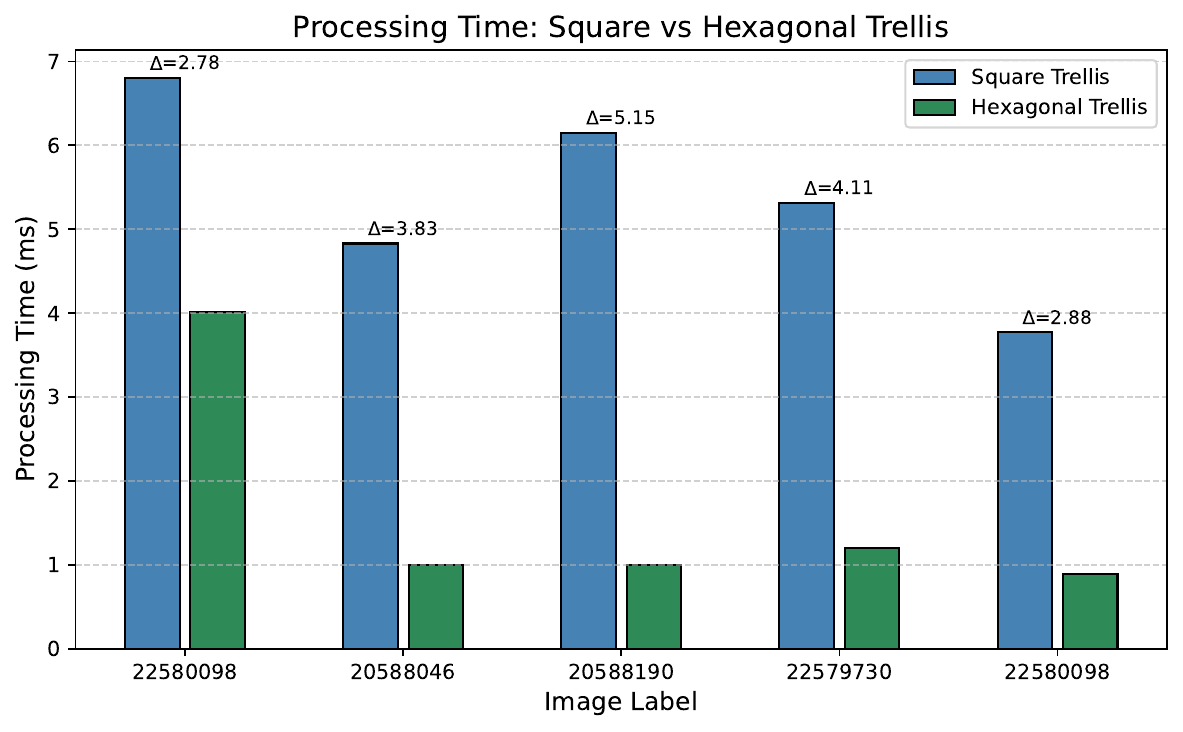}
\caption{Image processing time when implementing the square trellis-based method and NIVT method }
\label{fig:fig_19}
\end{figure*}

The results of the timing for each visualization of each image label are shown in Table 4 and also visualized in Figure \ref{fig:fig_18}. For example, image 22580099 took 6.80 ms to process via the conventional method and 4.016 ms using NIVT, meaning that NIVT saved an elapsed time of 3.984 sec. Also, 20588190 had the biggest time savings from NIVT, with a conventional time of 6.15 ms and NIVT time of 1.001 ms, resulting in 5.149 sec savings. Across all tested samples, the average difference stayed in the range of 2.88 to 5.15 sec. This analysis also shows that a greater proportion (and therefore the largest share relative to others) of the conventional method's overall time comes from the hexagonal domain filtering step. 

\section{Conclusion and Future Work}
The Virtual Hexagonal Trellis-Based Edge Detection method presents a novel approach for medical image analysis, particularly in breast cancer detection. By leveraging hexagonal pixel structures and optimized filter kernels, the method enhances edge continuity, reduces noise sensitivity, and improves computational efficiency. Future work includes deep-learning integration for automated edge refinement and real-time processing for clinical applications.
Hexagonal trellis-based edge detection techniques represent a promising advancement in the field of breast cancer lesion diagnosis. The reviewed studies highlight the potential benefits of hexagonal grids in enhancing edge detection accuracy and improving lesion classification performance. Future research should focus on developing efficient algorithms for grid conversion and optimizing computational processes to facilitate the widespread adoption of hexagonal trellis-based methods in clinical settings.

\section*{Statements and Declarations}

\subsection*{Funding}
Authors acknowledge the project Ongoing Research Funding program - Research Chairs (ORF-RC-2025-5300), King Saud University, Riyadh, Saudi Arabia.

\subsection*{Competing Interests}
The authors declare no conflicts of interest.

\subsection*{Author Contributions}
Shahid: initial draft, idea, writing; Rashid: idea improvements, writing;  all authors have read and agreed to the submitted version of the manuscript.

\subsection*{Data and Code Availability}
To support the research community, the processed version of the data (spectrogram images), along with all code and results, can be requested from the authors.

\bibliographystyle{unsrtnat}
\bibliography{sn-bibliography}

\end{document}